\documentclass[conference]{IEEEtran}
\IEEEoverridecommandlockouts

\usepackage{cite}
\usepackage{hyperref}
\usepackage{amsmath,amssymb,amsfonts}
\usepackage{algorithmic}
\usepackage{graphicx}
\usepackage{tabularx}
\usepackage{textcomp}
\usepackage{xcolor}
\usepackage{multirow}
\usepackage{booktabs}
\def\BibTeX{{\rm B\kern-.05em{\sc i\kern-.025em b}\kern-.08em
    T\kern-.1667em\lower.7ex\hbox{E}\kern-.125emX}}
\begin{document}

\title{Boundary-Guided Learning for Gene Expression Prediction in Spatial Transcriptomics}

\author{
\IEEEauthorblockN{Mingcheng Qu\textsuperscript{1}, Yuncong Wu\textsuperscript{2}, Donglin Di\textsuperscript{3}, Anyang Su\textsuperscript{4}, Tonghua Su\textsuperscript{1}, Yang Song\textsuperscript{5}, Lei Fan\textsuperscript{5}\IEEEauthorrefmark{2}}
\IEEEauthorblockA{\textsuperscript{1}\textit{Faculty of Computing}, \textit{Harbin Institute of Technology}, Harbin, China}
\IEEEauthorblockA{\textsuperscript{2}\textit{School of Astronautics}, \textit{Harbin Institute of Technology}, Harbin, China}
\IEEEauthorblockA{\textsuperscript{3}\textit{Space AI, Li Auto}, Beijing, China}
\IEEEauthorblockA{\textsuperscript{4}\textit{College of Software, Jilin University}, Changchun, China}
\IEEEauthorblockA{\textsuperscript{5}\textit{Computer Science and Engineering, UNSW}, Sydney, Australia}
\IEEEauthorblockA{\IEEEauthorrefmark{2}Corresponding author, Email: lei.fan1@unsw.edu.au}
}

\def\etal{\emph{et al. }}
\def\ie{\emph{i.e., }}
\def\eg{\emph{e.g., }}

\maketitle

\begin{abstract}

Spatial transcriptomics (ST) has emerged as an advanced technology that provides spatial context to gene expression. Recently, deep learning-based methods have shown the capability to predict gene expression from WSI data using ST data. Existing approaches typically extract features from images and the neighboring regions using pretrained models, and then develop methods to fuse this information to generate the final output. However, these methods often fail to account for the cellular structure similarity, cellular density and the interactions within the microenvironment.

In this paper, we propose a framework named BG-TRIPLEX, which leverages boundary information extracted from pathological images as guiding features to enhance gene expression prediction from WSIs. Specifically, our model consists of three branches: the spot, in-context and global branches. In the spot and in-context branches, boundary information, including edge and nuclei characteristics, is extracted using pretrained models. These boundary features guide the learning of cellular morphology and the characteristics of microenvironment through Multi-head Cross-Attention. Finally, these features are integrated with global features to predict the final output.

Extensive experiments were conducted on three public ST datasets and the Visium dataset. The results demonstrate that our BG-TRIPLEX consistently outperforms existing methods in terms of Pearson Correlation Coefficient (PCC). This method highlights the crucial role of boundary features in understanding the complex interactions between WSI and gene expression, offering a promising direction for future research. Codes are available at: https://github.com/WcloudC0416/BG-TRIPLEX.

\end{abstract}


\section{Introduction}

\begin{figure}
    \centering
    \includegraphics[width=0.5\textwidth]{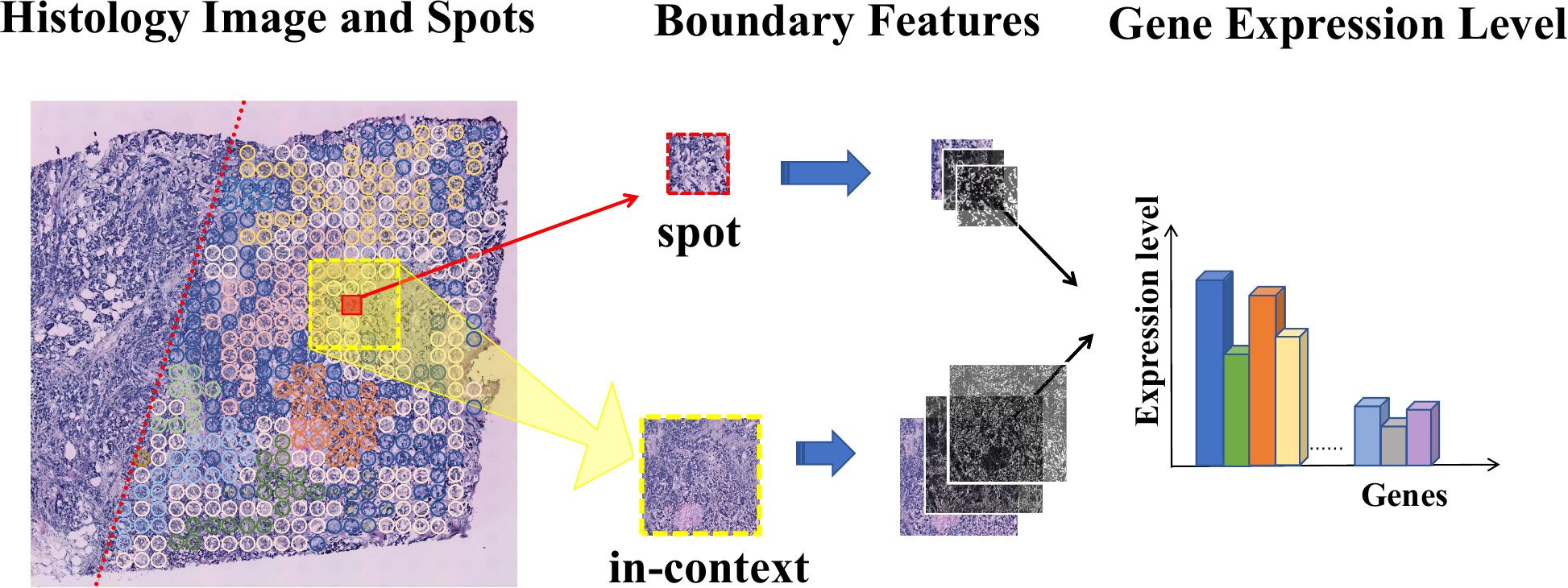}
    \caption{\textbf{Overview of our BG-TRIPLEX}. Our method extracts boundary information to guide the capture of cellular morphology and the characteristics of both the target spot and its in-context regions in histology images. These features are then fused to predict gene expression levels.}
    \label{fig:intro_method}
\end{figure}

Gene expression prediction plays a crucial role in cancer diagnosis and treatment \cite{apply_in_detection, apply_in_drug, apply_in_M&R}. Early detection techniques, such as bulk RNA sequencing \cite{bulk_RNA} and single-cell RNA sequencing (scRNA-seq) \cite{single_cell_RNA}, capture heterogeneity by isolating cells or analyzing gene expression in bulk tissues without spatial context. Recently, spatial transcriptomics (ST) \cite{st2016visualization, st2019, st2020, 2022st, 2024st} has emerged as a cutting-edge technology that provides spatial context to gene expression by capturing tissue expression through an array of spots. As illustrated in Fig.~\ref{fig:intro_method}, ST can be integrated with histological Whole Slide Imaging (WSI), where each ``spot'' on a tissue patch corresponds to a specific location from which gene expression is measured. The key advantage of ST is its ability to preserve the spatial organization of the tissue, which is crucial for understanding the complex interactions between cells within their native microenvironment \cite{st2016visualization}.

Deep learning-based computational methods have been shown to directly predict spatial gene expression from WSI data, which can significantly reduce the cost of ST by training models to map gene expression profiles to specific tissue regions. For example, ST-Net \cite{2020ST-Net} effectively predicts spatial gene expression by leveraging deep learning techniques to learn the spatial arrangement of gene expressions from histological images. In ST-Net, a pre-trained DenseNet \cite{DenseNet}, is employed for feature extraction to predict the expression levels of 250 highly expressed genes. HisToGene \cite{HistToGene} employs a Vision Transformer \cite{ViT} to capture spatial dependencies of spots in breast cancer, while Hist2ST \cite{Hist2ST} enhances prediction by incorporating both local and global spatial information. TRIPLEX \cite{2024TRIPLEX} captures features at individual spots, local context around these spots and global tissue organization by harnessing multiresolution features. Moreover, some methods \cite{2020ST-Net, ALLYOURNEED_singleSpot} primarily focus on image-level information relying on single spot images, while others \cite{HistToGene, Hist2ST, EGGN, BLEEP, THItoGene, 2024TRIPLEX} utilize information from multiple spots to capture broader spatial relationships. Despite their effectiveness, these approaches fail to account for the similarity and quantity of cellular structures in spatial expression prediction. Typically, spot images contain many similar cells and corresponding microenvironments, rich in boundary information such as cell density, nuclei morphology, and texture of different tissues. Explicitly incorporating this boundary information can effectively enhance cell type recognition, which is crucial for accurate gene expression prediction.

In this paper, we aim to utilize boundary information as guiding features to capture cellular characteristics of tissue structure and morphology, thereby enhancing the accuracy of gene expression prediction from WSIs. Specifically, we propose a Boundary-guided framework for spatial gene expression prediction, called BG-TRIPLEX. Our model is divided into three branches: the spot, in-context and global branches. Both the spot branch and in-context branch extract boundary information, including edge and nuclei details, from the spot image and its neighboring region respectively. This guiding mechanism is implemented using the Multi-Head Cross-Attention (MCA) \cite{MCA}, where boundary information serves as guiding features to capture relevant spatial details. Then, the spot and in-context features are obtained and fused with the global features, derived from the WSI using a global branch, to predict the final outcome. Through this guided learning approach, we effectively incorporate cell-level and nuclei-level information, resulting in highly accurate spatial gene expression predictions.

Our contributions can be summarized as follows:
\begin{itemize}
    \item We propose a novel framework, BG-TRIPLEX, which integrates features from different fields of view across three branches to enhance spatial gene expression prediction.
    \item We leverage boundary information, specifically edge and nuclei features, as guiding features to capture relevant spatial details through a Multi-Head Cross-Attention (MCA) for feature fusion.
    \item We conducted experiments on three ST datasets with WSI data to demonstrate the effectiveness of our model, and it produces competitive performance with a significant improvement of near 20.83\% over existing methods in the PCC metric.
\end{itemize}

\begin{figure*}
    \centering
    \includegraphics[width=1\textwidth]{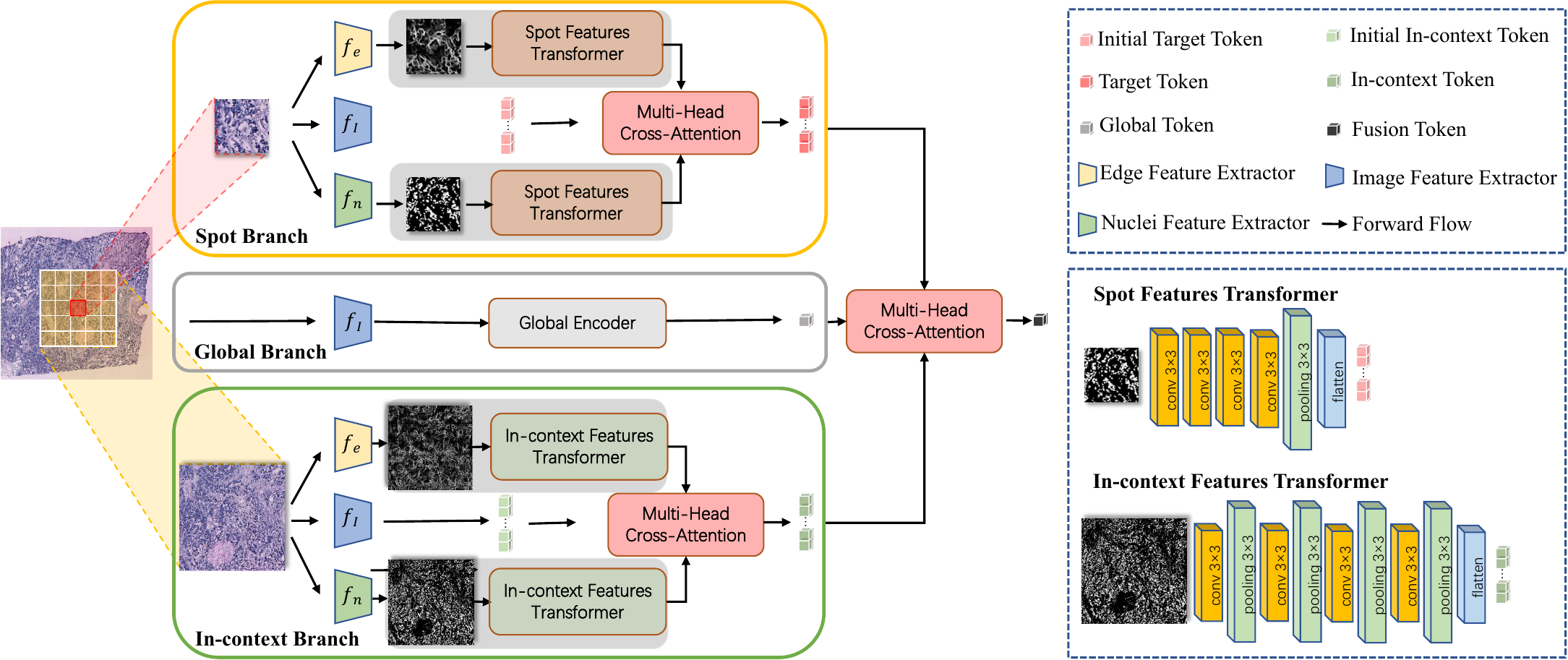}
    \caption{\textbf{The architecture of BG-TRIPLEX}. Our model consists of three main branches: Spot Branch, In-context Branch and Global Branch. In the spot and in-context branches, boundary features are extracted from the target patch and its in-context regions using pretrained models $f_e$ and $f_n$. These boundary features guide the capture of cellular characteristics through Multi-Head Cross-Attention. Finally, features from all three branches are fused to predict the gene expression.}
    \label{fig:framework}
\end{figure*}

\section{Related Work}

\subsection{Spatial Gene Expression Prediction}

Spatial Transcriptomics can be applied to WSIs by using the spots and corresponding gene expression data to train models capable of predicting spatial gene expression patterns. For example, ST-Net \cite{2020ST-Net} fine-tunes a pretrained model using histology images as input and gene expression data as labels. Following this, HisToGene \cite{HistToGene} leverages a Transformer model to learn the relationships among spots and is capable of predicting super-resolution gene expression. Hist2ST \cite{Hist2ST}, EGGN \cite{EGGN}, BLEEP \cite{BLEEP} and THItoGene \cite{THItoGene} have employed various techniques to extract features from histology image patches, including graph neural networks, exemplar learning, contrastive learning and capsule networks, respectively. Recently, TRIPLEX \cite{2024TRIPLEX} has utilized multiresolution features through a fusion strategy. Overall, previous methods have predominantly relied on local or neighboring information, often overlooking critical cellular details.

In our work, we extract boundary information from both the spot image and its in-context regions to guide feature learning, capturing cellular structures and spatial context for image feature representations.

\subsection{Boundary Information Extraction}

Boundary information is essential in feature learning by providing precise localization and detailed features, enabling models to accurately understand complex scenarios \cite{Boundary1, Boundary2, Boundary3}.

There are various models for extracting boundary features. HED \cite{HED} and DeepEdge \cite{DeepEdge} perform end-to-end edge detection using fully convolutional networks. RCF \cite{RCF} leverages rich convolutional features by combining multilayer convolutional information for edge detection. U-Net \cite{U-Net} employs an encoder-decoder architecture with skip connections to incorporate low-level details for object segmentation. MicroNet \cite{Micro-Net} enhances U-Net by introducing an improved architecture and weighted loss function, making it more robust in handling nuclei of different sizes. SAMS-Net \cite{SAMS-Net} addresses variations in staining by incorporating a weighted loss function that accounts for intensity variations in pathological images.

In this paper, we aim to use boundary information as guiding features to enhance spatial gene expression prediction. To achieve this, we employ PiDiNet \cite{PidiNet} and HoverNet \cite{Hover-Net} for extracting edge and nuclei information, taking advantage of their superior capabilities.

\section{Method}

\subsection{Preliminaries}
Whole Slide Images (WSIs) have giga-level resolutions that capture detailed images of tissue sections. Performing spatial gene inspection directly from WSI is challenging due to technical limitations and complexity. Spatial Transcriptomics (ST) achieves gene expression detection through a set of spatially arranged spots, where each ``spot'' represents a specific area on the tissue section. For a WSI with $N_s$ spots, each spot corresponds to an image patch $X \in \mathbb{R}^{H \times W \times 3} $, where $H$ and $W$ represent the height and width respectively.

Each spot also corresponds to measured gene expression profiles. Considering technical errors and signal-to-noise ratio \cite{2020ST-Net, 2024TRIPLEX}, we selected the top $K$ ($K$=250) most highly expressed genes in each dataset as labels for prediction and further identified 50 highly predictive genes during cross-validation to focus our predictive efforts. To normalize the data distribution and reduce heteroscedasticity, a logarithmic transformation was applied to the gene expression data, with a small constant added to all gene expression levels to avoid zero values. The processed gene expression \(G \in \mathbb{R}^{K}\) can thus be obtained as:
\begin{equation}
    G = log(G_o + 1) ,
\end{equation}
\noindent where \(G_o\) are the original gene expression levels respectively. For each spot position, a spot image X and its corresponding gene expression G can be paired. The aim is to train a model \(f\) to predict the expression levels, as defined:
\begin{equation}
G_i = f(\{X_i\}) \in \mathbb{R}^K,
\end{equation}
where $\{X_i\}$ denotes a set of images, indicating that the function \(f\) can take multiple images as input to produce the gene expression data \(G_i\).


\subsection{BG-TRIPLEX}
Unlike previous methods \cite{2020ST-Net, ALLYOURNEED_singleSpot} that primarily rely on spot image information, we aim to combine cellular, spot and neighboring information by explicitly utilizing boundary information as guiding features for extracting image features. Our proposed BG-TRIPLEX model comprises three branches: the spot branch, the in-context branch and the global branch, as shown in Fig. \ref{fig:framework}. We extract boundary information in the spot target and in-context branch and utilize Multi-Head Cross-Attention (MCA) \cite{MCA} to guide feature learning. Then, the features extracted from these three branches are then integrated using a feature fusion module to generate the final outcome.

\textbf{The Spot Branch.} This branch takes the target image $I_s \in \mathbb{R}^{H_s \times W_s \times 3}$ as input, where boundary information (\ie edge and nuclei details) and image features are extracted using three models: $f_e$, $f_n$ and $f_I$ respectively. We use pretrained ResNet18 \cite{he2016deep} as $f_I:\mathbb{R}^{H_s \times W_s \times 3}\rightarrow\mathbb{R}^{N_s \times D_s}$ to extract image features $\phi_I^{s}$, while pretrained PiDiNet \cite{PidiNet} and HoverNet \cite{Hover-Net} are used as $f_e:\mathbb{R}^{H_s \times W_s \times 3} \rightarrow \mathbb{R}^{N_s \times D_s}$ and $f_n:\mathbb{R}^{H_s \times W_s \times 3} \rightarrow \mathbb{R}^{N_s \times D_s}$ to extract edge features $\phi_{e}^s$ and nuclei features $\phi_{n}^s$. The spot feature transformer is used to match the image feature shape. 

\begin{figure}[t]
    \centering
    \includegraphics[width=0.45\textwidth]{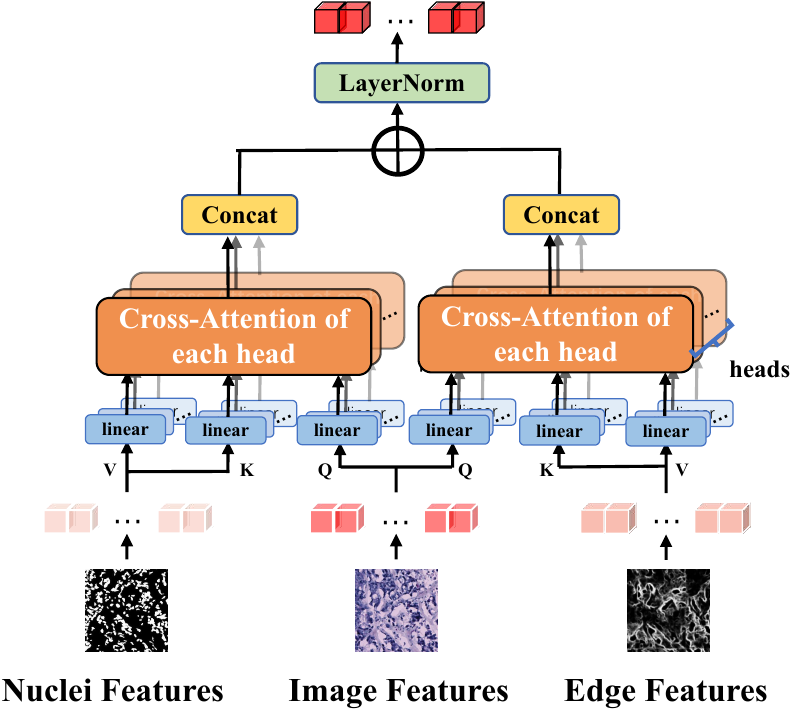}
    \caption{\textbf{The Multi-Head Cross-Attention for boundary-guided learning}. It involves generating the query $Q$ from the image features and the key $K$ and value $V$ from the boundary features. The boundary features guide the image features through cross-attention operations, after which the outputs are concatenated and passed through a LayerNorm layer.}
    \label{fig:MCA}
\end{figure}

Then, the boundary features $\phi_{e}^s$ and $\phi_{n}^s$ are utilized as references to guide the image features $\phi_I^{s}$ using MCA, as shown in Fig. \ref{fig:MCA}. Specifically, it includes two cross-attention operations between $\phi_I^{s}$ to $\phi_{e}^s$, $\phi_I^{s}$ to $\phi_{n}^s$. Taking an attention head $i$ as an example, $\phi_I^{s}$ is treated as the query and one of the boundary features as both the key and value. The attention output is calculated as:
\begin{equation}
\phi_{O}^{s} = \text{softmax}\left(\frac{(\mathbf{w}_q \phi_I^{s})(\mathbf{w}_k \phi_e^{s})^\top}{\sqrt{D_s}}\right) \mathbf{w}_v \phi_e^{s},    
\end{equation}
where $\mathbf{w}_q$, $\mathbf{w}_k$ and $\mathbf{w}_v$ are learnable projection matrices. The outputs of all heads are then concatenated to obtain the fused boundary features $\phi_{Ie}^{s}$ and $\phi_{In}^{s}$. They are further summed and fed into a LayerNorm to obtain the boundary-guided image features, as expressed below:
\begin{equation}
     \phi^{s} = \frac{(\phi_{Ie}^{s} + \phi_{In}^{s}) - \mu}{\sigma} \cdot \gamma + \beta ,
\end{equation}
where $\mu$ and $\sigma$ are the mean and the standard deviation of the input $\phi_{Ie}^{s}$ and $\phi_{In}^{s}$, $\gamma$ and $\beta$ are the scale and learnable shift parameters. Overall, the above process can be summarized as:
\begin{equation}
     \phi^{s} = f_{MCA}(\phi_{e}^{s},\phi_{I}^{s},\phi_{n}^{s}) ,
\end{equation}
where $f_{MCA}$ denotes the process of obtaining output $\phi^{s}$. Through this approach, the MCA mechanism captures intrinsic relationships between image features and different boundary-guided features, thereby enhancing the model's ability to learn detailed cellular characteristics.

\textbf{The In-context Branch.} Considering that the in-context regions are closely related to the current target image in terms of interactions and microenvironment \cite{2024TRIPLEX}, we further incorporate an in-context branch to extract neighboring regions surrounding the target image $I_S$. Specifically, we extract $D\times D$ patches with the same image size as $I_S$, and the in-context image is defined as $I_{in} \in \mathbb{R}^{H_s\cdot D \times W_s\cdot D \times 3} $. 

Similar to the spot branch, we extract image features and boundary features for the in-context image $I_{in}$ using three models $f_I$, $f_e$ and $f_n$. This results in the in-context image features $\phi_I^{in} \in \mathbb{R}^{N_{in} \times D_{in}}$, edge features $\phi_e^{in} \in \mathbb{R}^{N_{in} \times D_{in}}$ and nuclei features $\phi_n^{in} \in \mathbb{R}^{N_{in} \times D_{in}}$. An in-context feature transformer is then employed to align the feature shapes. Subsequently, the in-context boundary features $\phi_e^{in}$ and $\phi_n^{in}$ are used to guide the in-context features $\phi_I^{in}$ through the MCA as defined: 
\begin{equation}
     \phi^{in} = f_{MCA}(\phi_{e}^{in},\phi_{I}^{in},\phi_{n}^{in}),
\end{equation}
By incorporating boundary information into the in-context branch, BG-TRIPLEX gains the capability to better capture and integrate crucial spatial and morphological details from the neighboring regions of the target spot. This enhanced representation of the microenvironment surrounding the spot is critical for accurately modeling the complex interactions that influence gene expression.

\textbf{The Global Branch.}
This branch processes all spot patches within the WSI to extract a global view of the entire WSI, thereby assisting the model in comprehending the overall structure and layout of the tissue. 
Given that the foreground of WSIs can vary in irregular shapes and size, we follow the previous studies \cite{2024TRIPLEX} and adopt the Atypical Position Encoding Generator (APEG) to reorganize the spatial contexts according to their relative positions. APEG effectively encodes the absolute positional information of each spot patch into the model, which is crucial for accurately capturing the spatial distribution within the WSI.

Specifically, the global branch takes $I^g \in \mathbb{R}^{H_g \times W_g \times 3}$ as input and leverages $f_I$ to extract features $\phi_I^{g} \in \mathbb{R}^{N_{g} \times D_{g}}$ from the WSI. Incorporating the global branch not only enhances the model's understanding of each cell’s contextual location within the tissue but also minimizes redundant feature extraction for individual spots, thus enhancing the model’s overall efficiency. 

\textbf{Branch Fusion.} After obtaining the output features $\phi^{s}$, $\phi^{in}$ and $\phi_I^{g}$ from the spot, in-context and global branches respectively, we integrate them to produce the final fused features for the spatial gene expression prediction. The fusion process is carried out using the MCA mechanism, as expressed:
\begin{equation}
    \phi^{fusion} = f_{MCA}(\phi^{s},\phi_{I}^{g},\phi^{in}) , 
\end{equation}

By integrating features from three branches with multi-scale fields of view, the model captures a broader range of factors influencing gene expression. This fusion allows the model to consider multiple perspectives, ranging from localized cellular information to tissue context, thereby providing a more holistic understanding of spatial gene expression patterns.

\textbf{Motivation.}
Our approach leverages features directly related to cellular morphology, including nuclei information and edge information, as guiding features for extracting other features. This method enhances the model's ability to capture and analyze the detailed morphology of cells within WSIs. Nuclei information provides critical insights into the internal structure of cells, essential for distinguishing various cell types and understanding their functional roles. Meanwhile, edge information helps delineate cell boundaries and their interactions with neighboring cells. By incorporating these morphology-related features into the feature extraction process, the model achieves greater accuracy in recognizing and classifying cells based on their morphology, leading to more precise and reliable predictions of spatial gene expression.

\begin{table*}[t]
    \caption{Performance Comparison of Different Models on three ST Datasets.}
    \centering
    \resizebox{\textwidth}{!}{%
    \begin{tabular}{|l|c|c|c|c|c|c|c|c|c|c|c|c|}
        \toprule
        \midrule
        \multicolumn{1}{|c}{\multirow{2}[0]{*}{Models}} & \multicolumn{3}{|c|}{\textbf{HER2ST}} & \multicolumn{3}{c|}{\textbf{STNet}} & \multicolumn{3}{c|}{\textbf{SKin}} \\
        \cmidrule{2-4} \cmidrule{5-7} \cmidrule{8-10}
        & \textbf{MSE ($\downarrow$)} & \textbf{PCC(M) ($\uparrow$)} & \textbf{PCC(H) ($\uparrow$)} & \textbf{MSE ($\downarrow$)} & \textbf{PCC(M) ($\uparrow$)} & \textbf{PCC(H) ($\uparrow$)} & \textbf{MSE ($\downarrow$)} & \textbf{PCC(M) ($\uparrow$)} & \textbf{PCC(H) ($\uparrow$)} \\
        \hline
        ST-Net~\cite{2020ST-Net} & 0.260 $\pm$ 0.04 & 0.194 $\pm$ 0.11 & 0.345 $\pm$ 0.16 & 0.209 $\pm$ 0.02 & 0.116 $\pm$ 0.06 & 0.223 $\pm$ 0.10 & 0.294 $\pm$ 0.07 & 0.274 $\pm$ 0.08 & 0.382 $\pm$ 0.08 \\
        EGN~\cite{EGGN} & 0.241 $\pm$ 0.06 & 0.197 $\pm$ 0.11 & 0.328 $\pm$ 0.17 & 0.192 $\pm$ 0.02 & 0.111 $\pm$ 0.05 & 0.203 $\pm$ 0.09 & 0.281 $\pm$ 0.08 & 0.281 $\pm$ 0.06 & 0.388 $\pm$ 0.06 \\
        BLEEP~\cite{BLEEP} & 0.277 $\pm$ 0.05 & 0.151 $\pm$ 0.11 & 0.277 $\pm$ 0.10 & 0.235 $\pm$ 0.02 & 0.145 $\pm$ 0.05 & 0.193 $\pm$ 0.10 & 0.297 $\pm$ 0.08 & 0.269 $\pm$ 0.07 & 0.396 $\pm$ 0.08 \\
        HistoGene~\cite{HistToGene} & 0.248 $\pm$ 0.05 & 0.178 $\pm$ 0.13 & 0.308 $\pm$ 0.14 & 0.199 $\pm$ 0.03 & 0.122 $\pm$ 0.10 & 0.201 $\pm$ 0.09 & 0.294 $\pm$ 0.07 & 0.312 $\pm$ 0.07 & 0.406 $\pm$ 0.10 \\
        His2ST~\cite{Hist2ST} & 0.285 $\pm$ 0.08 & 0.118 $\pm$ 0.10 & 0.258 $\pm$ 0.11 & 0.181 $\pm$ 0.07 & 0.144 $\pm$ 0.07 & 0.199 $\pm$ 0.07 & 0.291 $\pm$ 0.16 & 0.174 $\pm$ 0.16 & 0.353 $\pm$ 0.07 \\
        TRIPLEX~\cite{2024TRIPLEX} & \textbf{0.228$\pm$ 0.07} & 0.314 $\pm$ 0.14 & 0.497 $\pm$ 0.17 & 0.202 $\pm$ 0.07 & 0.206 $\pm$ 0.07 & 0.352 $\pm$ 0.10 & 0.268 $\pm$ 0.09 & 0.374 $\pm$ 0.07 & 0.490 $\pm$ 0.07 \\
        \midrule
        \textbf{TG-TRIPLEX} & 0.277 $\pm$ 0.01 & \textbf{0.366 $\pm$ 0.02} & \textbf{0.605 $\pm$ 0.02}  & \textbf{0.1338 $\pm$ 0.01} & \textbf{0.4978 $\pm$ 0.02} & \textbf{0.7234 $\pm$ 0.04} & \textbf{0.128 $\pm$ 0.01} & \textbf{0.655 $\pm$ 0.01} & \textbf{0.752 $\pm$ 0.01} \\
        \midrule
        \bottomrule
    \end{tabular}}
    \label{table: cross comparison}
\end{table*}

\subsection{Training loss} 
\label{Loss}

Our BG-TRIPLEX integrates features from three branches, each providing a different field-of-view. The optimization objective is composed of two parts $mathcal{L}_{f}$ and $mathcal{L}_{s}$. Specifically, $mathcal{L}_{f}$ corresponds to minimizing the Mean Squared Error (MSE) between the true values and the final fused outcome, as defined by:
\begin{equation}
    \mathcal{L}_{f} = \frac{1}{k} \sum_{j}^k \left( p_f^j - g^j \right)^2 ,
\end{equation}
where $g^j$ and $p_f^j$ represent the label and the predicted gene expression using the fused features for the $j$-th gene ($j \in \{1,\cdots, k\}$). In addition, the loss $\mathcal{L}_{s}$ is calculated for each individual branch by comparing its prediction to both the true label and the final fused prediction, as defined by:
\begin{equation}
    \mathcal{L}_{s}^i = (1 - \lambda) \frac{1}{k} \sum_{j}^k \left( p_i^j - g^j\right)^2 + \lambda \frac{1}{k} \sum_{j}^k \left( p_i^j - p_f^j \right)^2 ,
\end{equation}
where $p_i^j$ denotes the prediction of the $i$-th branch with $i$ indicating the index of the spot, in-context or global branches, and $\lambda$ is a hyperparameter to balance different losses. The overall optimization objective $\mathcal{L}_{total}$ is defined as:
\begin{equation}
    \mathcal{L}_{toal} = \sum_{i}\mathcal{L}_{b}^i + \mathcal{L}_{f} ,
\end{equation}

\section{Experiments}

\subsection{Datasets}

We conducted experiments on three publicly available datasets: HER2ST \cite{her2st}, STNet \cite{2020ST-Net} and Skin \cite{skin} datasets. 

\textbf{HER2ST dataset} is part of the Human Epidermal Growth Factor Receptor 2 (HER2) ST studies, focusing on breast cancer tissue sections diagnosed with invasive ductal carcinoma. It consists of spatially resolved transcriptomes encompassing 2,518 spots, with expression data for 17,651 genes. 

\textbf{STNet dataset} provides WSIs and corresponding gene expression profiles for various tissue types, offering a comprehensive view of the tissue microenvironment. It includes 10,000 spots across multiple tissue types, with each spot measuring the expression of over 20,000 genes. 

\textbf{Skin dataset} provides ST for skin tissue sections, including samples from both healthy and diseased states. It comprises spatially resolved gene expression data with approximately 5,000 spots, each measuring the expression levels of around 18,000 genes.

\subsection{Implementation Details}

We trained all models using the PyTorch platform. For data preprocessing, for each pathological tissue slice image in the dataset, boundary information was generated using pretrained models. Specifically, for the edge features, we generated feature maps for each pathological tissue slice image in the dataset using PiDiNet \cite{PidiNet}, which was pretrained on the BSDS500 dataset \cite{arbelaez2010contour}. We obtained cell nucleus segmentation maps using pretrained HoverNet \cite{Hover-Net}. Both types of generated boundary features were grayscale and had the same dimensions as the corresponding original images. 

For the target branch, the spot image was extracted and resized to a fixed size of $224 \times 224$ pixels, while the input for the in-context branch was resized to $1120 \times 1120$ pixels when $D$ was set to 5, indicating that $5 \times 5$ patches surrounding the target spot were used. The model was optimized using the Adam optimizer with an initial learning rate of 0.0001, and the learning rate was dynamically adjusted with a step size of 50 and a decay rate of 0.9. The batch size during training was 12. The number of training epochs was set to 20. The $\lambda$ was set to 0.3.

\begin{table*}[t]
\centering
\caption{Performance Comparison of Different Models on 10X Visium Datasets.}
\begin{tabular}{|l|ccc|ccc|ccc|}
\toprule
\midrule
\multicolumn{1}{|c}{\multirow{2}[0]{*}{Models}} & \multicolumn{3}{|c|}{10X Visium-1} & \multicolumn{3}{c|}{10X Visium-2} & \multicolumn{3}{c|}{10X Visium-3} \\
\cmidrule{2-4} \cmidrule{5-7} \cmidrule{8-10}
                       & MSE ($\downarrow$)  & PCC(M) ($\uparrow$) & PCC(H) ($\uparrow$) & MSE ($\downarrow$)  & PCC(M) ($\uparrow$) & PCC(H) ($\uparrow$) & MSE ($\downarrow$)  & PCC(M) ($\uparrow$) & PCC(H) ($\uparrow$) \\ \hline
ST-Net \cite{2020ST-Net}         & 0.423 & -0.026 & -0.000 & 0.395 & 0.091  & 0.193  & 0.424 & -0.032 & 0.008  \\
EGN \cite{EGGN}          & 0.421 & 0.003  & 0.024  & 0.328 & 0.102  & 0.157  & 0.303 & 0.106  & 0.220  \\
BLEEP \cite{BLEEP}        & 0.367 & 0.106  & 0.221  & 0.289 & 0.104  & 0.260  & 0.298 & 0.114  & 0.229  \\

TRIPLEX  \cite{2024TRIPLEX}           & \textbf{0.351}  & \textbf{0.136} & 0.241 & \textbf{0.282}  & \textbf{0.155} & 0.356 & \textbf{0.285}  & \textbf{0.118} & \textbf{0.282} \\ 
\midrule
\textbf{BG-TRIPLEX}       & 0.366  & 0.074 & \textbf{0.342} & 0.328  & 0.115 & \textbf{0.397}  & 0.325 & 0.036 & 0.275 \\
\midrule
\bottomrule
\end{tabular}
\label{table: general performance}
\end{table*}

\subsection{Evaluation Metrics}

In our experiment, we selected Mean Squared Error (MSE) and Pearson Correlation Coefficient (PCC) as evaluation metrics, as they provide a comprehensive assessment of the model's performance from different perspectives. 

\textbf{MSE} primarily evaluates the magnitude of the error between predicted values and true values, with a smaller MSE indicating that the model's predictions are closer to the true values. The MSE is calculated as follows:
\begin{equation}
\text{MSE} = \frac{1}{k \times N_s}\sum_{i=1}^{k}\sum_{j=1}^{n}(\hat{y}_{i,j} - y_{i,j})^2 ,
\end{equation}
where $y_{i,j}$ represents the $j$-th actual value for the $i$-th gene, $\hat{y}_{i,j}$ denotes the corresponding predicted value, and $k$ and $N_s$ are numbers of genes and spots, respectively. 

\textbf{PCC} measures the linear correlation between predicted values and true values, reflecting the consistency and strength of the relationship between the predictions and the actual data. This metric is particularly important for assessing the model's ability to capture underlying patterns and trends in the data. PCC values range from -1 to 1, with values closer to 1 indicating a stronger correlation between the predicted and actual values. The PCC is calculated as follows:
\begin{equation}
\text{PCC} = \frac{\sum_{i=1}^{k}\sum_{j=1}^{N_s}(\hat{y}_{i,j} - \bar{\hat{y}}_{.,j})(y_{i,j} - \bar{y}_{.,j})}{\sqrt{\sum_{i=1}^{k}\sum_{j=1}^{N_s}(\hat{y}_{i,j} - \bar{\hat{y}}_{.,j})^2} \sqrt{\sum_{i=1}^{k}\sum_{j=1}^{N_s}(y_{i,j} - \bar{y}_{.,j})^2}} .
\end{equation}
where $y_{i,j}$ represents the $j$-th true value for the $i$-th gene, and $\hat{y}_{i,j}$ denotes the corresponding predicted value. $\bar{\hat{y}}_{.,j}=\frac{1}{k}\sum_{i}{y_{i,j}}$ and $\bar{y}_{.,j}=\frac{1}{k}\sum_{i}{y_{i,j}}$ denote the mean of all predicted values and true values respectively.

We calculated the average PCC for all gene expressions, denoted as PCC(M), and the average PCC for the top 50 highly expressed genes, denoted as PCC(H), to assess highly expressed genes. By using MSE, PCC(M) and PCC(H), we can comprehensively evaluate the model's performance from the perspectives of error magnitude and prediction correlation.

\subsection{Results and Discussion}

\textbf{Cross-validation performance on ST datasets.}
We conducted experiments on the HER2ST, STNet, and Skin datasets to compare our BG-TRIPLEX model with advanced methods, including ST-Net \cite{2020ST-Net}, EGN \cite{EGGN}, BLEEP \cite{BLEEP}, HisToGene \cite{HistToGene}, Hist2ST \cite{Hist2ST}, and TRIPLEX \cite{2024TRIPLEX}, as shown in Table \ref{table: cross comparison}. Our BG-TRIPLEX achieved almost overall improvement in both MSE and PCC across all datasets. It obtained the lowest MSE on the STNet and Skin datasets. Notably, the PCC values showed significant improvements across all three datasets, particularly in the Skin dataset, where PCC(M) and PCC(H) reached 0.655 and 0.752 respectively, reflecting improvements of 0.281 and 0.262. These performance metrics demonstrate BG-TRIPLEX's ability to predict gene expression with lower errors and higher correlations.

\textbf{Generalization performance on Visium data.}
We further tested our model on unseen Visium samples to evaluate its generalization performance. The results showed in a significant improvement compared to previous models, demonstrating the robustness and applicability of our approach. As shown in Table~\ref{table: general performance}, specifically, while the MSE exhibited moderate performance across the three 10X Visium datasets, it remained a relatively stable value. The PCC (M) showed strong results on the 10X Visium-2 dataset, and PCC (H) performed well across all three 10X Visium datasets, even surpassing previously optimal models. This indicates that BG-TRIPLEX possesses robust capabilities when handling highly expressed genes.

\begin{figure}[t]
    \centering
    \includegraphics[width=0.48\textwidth]{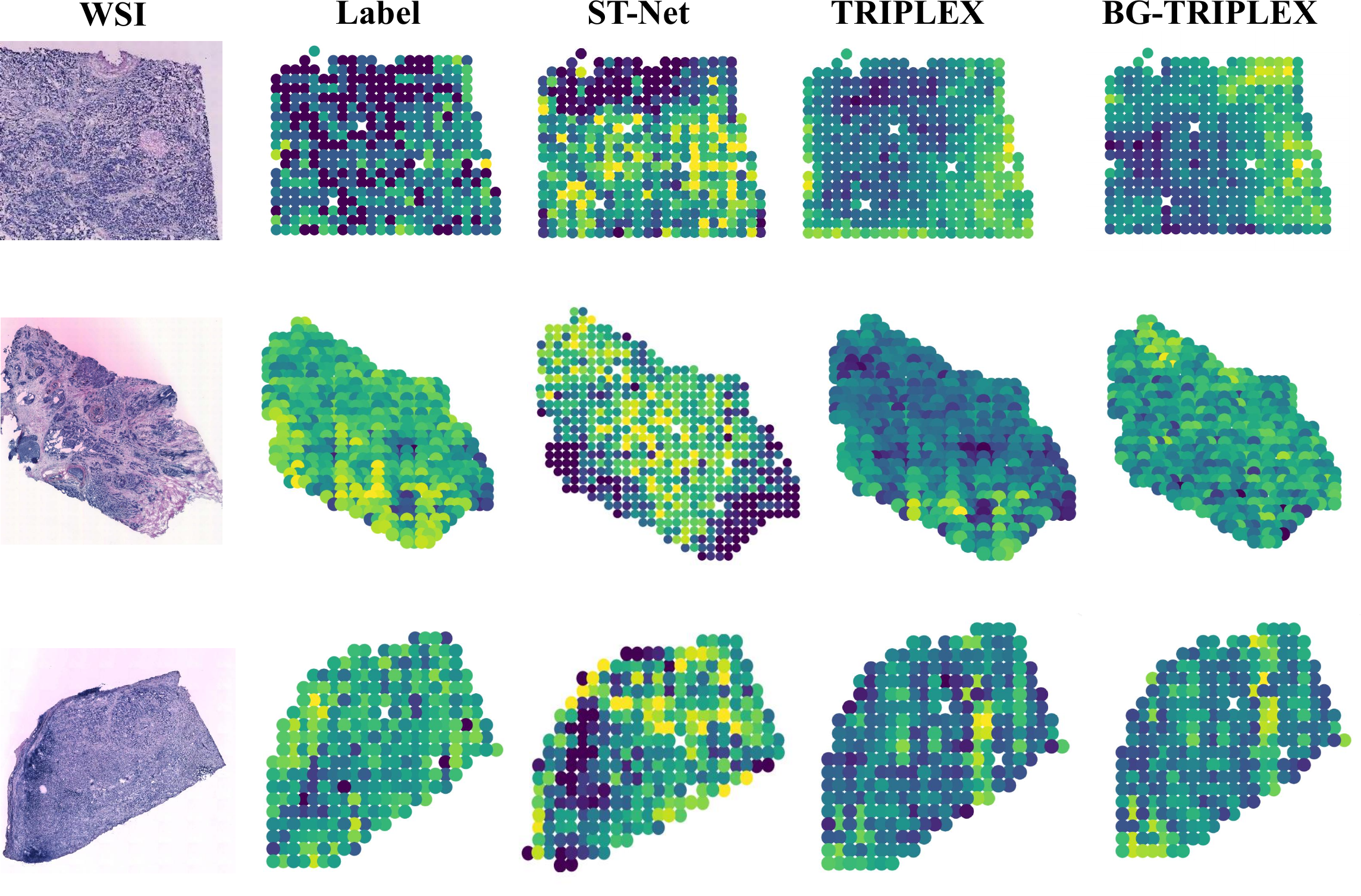}
    \caption{\textbf{Prediction visualizations on the STNet dataset}. From left to right: raw WSI, label of gene expression, ST-Net, TRIPLEX and our BG-TRIPLEX.}
    \label{fig:visualization}
\end{figure}

\textbf{Gene prediction visualization.} We presented visualizations of predicted gene values alongside their true labels for each model on the STNet dataset. As illustrated in Fig.~\ref{fig:visualization}, we displayed highly expressed prediction maps from BG-TRIPLEX in the final column and compared them with the ground truth, ST-Net \cite{2020ST-Net} and TRIPLEX \cite{2024TRIPLEX} predictions. We observed that our BG-TRIPLEX exhibits a high degree of visual consistency with the true gene expression. This suggests that BG-TRIPLEX is effective in accurately predicting gene expression patterns associated with breast cancer markers, thereby enhancing the reliability of pathological assessments.

\subsection{Ablation Studies}

We conducted extensive experiments to verify the effectiveness of each component of our model on the HER2ST dataset.

\textbf{Different branches.}
We evaluated the impact of each branch in BG-TRIPLEX on the accuracy of gene expression prediction, as shown in Table~\ref{table: Branches}. It was evident that the removal of any branch from BG-TRIPLEX led to a decrease in model performance. In particular, the absence of the in-context branch resulted in a decrease in PCC(M) and PCC(H) by 0.077 and 0.087, respectively. The removal of all three branches impacted the model's performance to varying degrees, with the most significant decrease observed in the two branches containing boundary information, \ie the spot and the in-context branches. This validated the effectiveness of each branch in our BG-TRIPLEX model.

\begin{table}[t]
\centering
\caption{Ablation study of Different Branches.}
\begin{tabular}{|l|c|c|c|}
\toprule
\midrule
Configuration & MSE ($\downarrow$) & PCC(M) ($\uparrow$) & PCC(H) ($\uparrow$)\\ \hline
\textbf{w/o.} Spot Branch      & 0.3343   & 0.3172      & 0.5635      \\
\textbf{w/o.} In-context Branch    & 0.2897   & 0.2889     & 0.5177     \\
\textbf{w/o.} Global Branch       & 0.3087   & 0.3041      & 0.5584      \\
\midrule
BG-TRIPLEX    & \textbf{0.277}   & \textbf{0.366}      & \textbf{0.605}      \\ 
\midrule
\bottomrule
\end{tabular}
\label{table: Branches}
\end{table}

\begin{table}[b]
\centering
\caption{Contribution of Boundary Features.}
\resizebox{0.5\textwidth}{!}{
\begin{tabular}{|l|c|c|c|}
\toprule
\midrule
Configuration & MSE ($\downarrow$)& PCC(M) ($\uparrow$) & PCC(H) ($\uparrow$)\\ \hline
\textbf{w/o.} Edge in Spot       & 0.2872   & 0.2731     & 0.5409     \\
\textbf{w/o.} Nuclei in Spot       & 0.2852   & 0.2719      & 0.5359     \\
\textbf{w/o.} Edge in In-context     & \textbf{0.2637}   & 0.2885     & 0.5493      \\
\textbf{w/o.} Nuclei in In-context       & 0.2761   & 0.2848  & 0.5428       \\
\midrule
BG-TRIPLEX    & 0.277   &  \textbf{0.366}     &  \textbf{0.605}   \\ 
\midrule
\bottomrule
\end{tabular}
}
\label{table: Boundary}
\end{table}

\textbf{Boundary features.}
We evaluated the impact of guiding information in different branches on the accuracy of gene expression prediction, as shown in Table~\ref{table: Boundary}. The experimental results demonstrated that each component of BG-TRIPLEX, particularly the edge and nuclei information in the spot and in-context branches, was crucial for maintaining performance. Removing edge information from the spot branch resulted in a significant decrease in both PCC(M) (from 0.366 to 0.2731) and PCC(H) (from 0.605 to 0.5409), highlighting its critical role. Similarly, the removal of nuclei information in the spot branch led to reduced performance, though the effect on PCC(H) was less pronounced. In the in-context branch, the most significant performance drop occurred when nuclei information was removed, with PCC(M) decreasing from 0.366 to 0.2848 and PCC(H) from 0.605 to 0.5428. These results validate the importance of incorporating both edge and nuclei information, particularly in the spot branch, for accurate gene expression prediction.

\textbf{Guiding methods.}
We further compared different guiding methods for boundary information, highlighting the effectiveness of MCA in our model. As shown in Table~\ref{table：GuidedMethod}, the MCA method performed the best overall, particularly in PCC(M) and PCC(H), achieving 0.366 and 0.605 respectively. This indicates its superior ability to capture complex gene expression patterns. While the summation operation excelled at reducing MSE (0.2707), it did not enhance gene expression correlation as effectively as MCA. The concatenation method performed the worst, with the highest MSE and the lowest PCC scores, indicating its weaker effectiveness in guiding feature integration.

\begin{table}[t]
\centering
\caption{Contribution of Guiding Method.}
\begin{tabular}{|l|c|c|c|}
\toprule
\midrule
Configuration & MSE ($\downarrow$)& PCC(M) ($\uparrow$)& PCC(H) ($\uparrow$)\\ \hline
Summation      & \textbf{0.2707}   & 0.3484     & 0.5972     \\
Concat    & 0.2875   & 0.2467     & 0.5166      \\
MCA       & 0.277   & \textbf{0.366}     & \textbf{0.605} \\ 
\midrule
\bottomrule
\end{tabular}
\label{table：GuidedMethod}
\end{table}

\section{Conclusion}
In this paper, we introduce a framework, named BG-TRIPLEX, that leverages boundary information, specifically edge and nuclei features, as guiding features to enhance the spatial gene expression prediction from WSIs. Our BG-TRIPLEX model comprises three branches: the spot, in-context and global branches. In the spot and in-context branches, edge and nuclei features are extracted using pretrained PiDiNet and HoverNet models respectively. By employing Multi-head Cross-Attention, these features guide the learning of cellular morphology and microenvironment characteristics from image features. Finally, features from all three branches are fused to predict the final output. 

Extensive experiments demonstrate that integrating boundary information significantly improves the model's performance. Tests on ST datasets and Visium Data show that BG-TRIPLEX not only excels in spatial gene expression prediction but also highlights the crucial role of boundary features in understanding the complex interactions between WSI and gene expression. This innovation provides a new approach for more precise and biologically meaningful interpretations of pathological images in gene expression analysis.


\bibliographystyle{IEEEtran} 
\bibliography{BG}

\end{document}